\documentclass[runningheads]{llncs}
\usepackage{subcaption}
\usepackage{graphicx}
\usepackage{subcaption}
\usepackage{amsfonts}
\usepackage{t1enc}

\usepackage{amsmath,amssymb} 
\usepackage{arydshln}
\usepackage{algorithm2e}
\begin{document}
\title{Deep Recurrent Neural Network\\for Multi-target Filtering}
%
%
\author{Mehryar~Emambakhsh\and
Alessandro~Bay\and
Eduard~Vazquez}
\authorrunning{M. Emambakhsh et al.}
%
\institute{Cortexica Vision Systems, London, UK 
\email{\{mehryar.emambakhsh,~alessandro.bay,~eduard.vazquez\}@cortexica.com}}
\maketitle              
\begin{abstract}
This paper addresses the problem of fixed motion and measurement models for multi-target filtering using an adaptive learning framework. This is performed by defining target tuples with random finite set terminology and utilisation of recurrent neural networks with a long short-term memory architecture. A novel data association algorithm compatible with the predicted tracklet tuples is proposed, enabling the update of occluded targets, in addition to assigning birth, survival and death of targets. The algorithm is evaluated over a commonly used filtering simulation scenario, with highly promising results.\footnote{\url{https://github.com/mehryaragha/MTF}}

\end{abstract}

\section{Introduction}
Multi-target filtering consists of automatically excluding clutter from (usually unlabelled) input data sequences. It has numerous applications in denoising spatio-temporal data, object detection and recognition, tracking, and data-, object- and track-level sensor fusion \cite{Vo:2005,Vo:2014,Fantacci:2018}. It is also frequently used as part of various military applications (target recognition), automation pipelines, autonomous vehicles, and localisation and occupancy-grid mapping (using optical, e.g.\ stereo camera or LiDAR, or radar sensors) for robotics.
Particularly for a tracking problem, the performance of a multi-target tracking algorithm heavily relies on its filtering step. A robust filtering algorithm
is capable of considering occlusions, probability of detection, possibility of birth and spawn of targets, and incorporating clutter densities. 
An accurate filtering algorithm can improve the lifespan of the generated tracklets and the localisation of targets.
Target state estimation by a filtering algorithm can be achieved by considering prior motion and measurement models. In a Bayesian filtering framework, motion models are used to predict the location of the target at the next time step, while measurement models are used to map predictions to the measurement space to perform correction (the update step).
Due to the high complexity of a target motion, the use of fixed models, however, may not result in satisfactory outputs and therefore, can deteriorate the filtering performance.

%

Considering this challenge and inspired by the recurrent neural network tracker proposed in \cite{Milan:2017} and random finite sets (RFS) multi-target filtering paradigm \cite{Mahler:2003,Mahler:2007,Vo:2006}, in this work, we propose a novel algorithm to perform multi-target filtering while simultaneously learning the motion model. To this end, a long short-term memory (LSTM,{\cite{hochreiter1997}}) recurrent neural network (RNN) architecture is defined over sets of target tuples and trained {online} using the incoming data sequences. The prediction step is performed by applying the trained LSTM network to data patch generated from targets. This LSTM network (which is then trained using the new updated states) gradually learns a global transferable motion model of the detected targets.
After obtaining the measurements, we use a novel data association algorithm, which is compatible with the generated tracklet tuples to assign survivals, deaths and births of new targets. The updated patches for each target are then used to train the LSTM model in the next time steps.
To evaluate our algorithm we have designed a multi-target simulation scenario. During the simulation, by increasing the clutter (false positive rate) intensity, we evaluate the filtering robustness. 
This work which is one of the first papers addressing the multi-target filtering task with recurrent neural networks shows remarkable potential, while outperforming well-known multi-target filtering approaches.

This paper is organised as follows: First, related work in the literature is explained in Section~\ref{sec:relwork} and the overall pipeline is briefly explained in Section~\ref{sec:overall}. Incorporation of recurrent neural networks for motion modelling is explained in Section~\ref{sec:lstmMerge}. Tracklet tuples and data association is explained in Section~\ref{sec:DataAs}. Section~\ref{sec:expRes} is dedicated to the experimental results. We conclude the paper in Section~\ref{sec:conc}, giving future working directions.

\subsection{Scientific contribution}

The proposed algorithm is capable of filtering multiple targets, with non-linear motion and non-Gaussian error models. Unlike \cite{Reuter:2014,Vo:2014}, no prior motion modelling is performed and mapping from state to observation space is learned from the incoming data sequence. {Moreover,} unlike \cite{Milan:2017} the proposed algorithm is being trained online and does not rely on a separate training {phase}.
Since the predicted targets are being concatenated over time within the target tuples, higher Markov order is preserved, enabling longer term target state memorisation. Compared with RFS multi-target filtering algorithms \cite{Reuter:2014,Vo:2014}, there are significantly fewer number of hyper parameters for the proposed algorithm. For example, \cite{Vo:2006} requires hyper parameters to perform pruning, merge and truncation of the output density function, in addition to clutter distribution, survival and detection probabilities. To the best of our knowledge, this is one of the first papers addressing the multi-target filtering task with recurrent neural networks, particularly without the need of pre-training the network.

\section{Related work}
\label{sec:relwork}

\noindent \textbf{{Model-based approaches:}}
In a Bayesian formulation of a multi-target filtering, the goal is to estimate the (hidden) target state $x_{k}$ at the $k^{th}$ time step, from a set of observations in the previous time steps $z_{1:k}$, {i.e.},
\begin{equation}
p_k(x_k|z_{1:k}) = \frac{g_k(z_k|x_k)p_{k|k-1}(x_k|z_{1:k-1})}{\int g_k(z_k|x)p_{k|k-1}(x|z_{1:k-1})dx},
\label{eq:bayes1}
\end{equation} 
\noindent where $g_k(z_k|x_k)$ and $p_{k|k-1}(x_k|z_{1:k-1})$ are the likelihood and transition densities, respectively. From (\ref{eq:bayes1}), it is clear that this is an recursive problem. The state estimation at the $k^{th}$ iteration is usually obtained by Maximum A Posteriori (MAP) criterion.
Kalman filter is arguably the most popular online filtering approach. It assumes linear motion models with Gaussian distributions for both of the prediction and update steps. Using the Taylor series expansion and deterministic approximation of non-Gaussian distributions, non-linearity and non-Gaussian behaviour are addressed by Extended and Unscented Kalman Filters {(EKF, UKF)}, respectively. Using the importance sampling principle, particle filters are also used to estimate the likelihood and posterior densities. Particle filters are one of the most widely used multi-target filtering algorithms capable of addressing non-linearity and non-Gaussian motions \cite{Vo:2003,Vo:2005}.

Mahler proposed random finite sets (RFS) formulation for multi-target filtering \cite{Mahler:2003}. RFS provides an encapsulated formulation for multi-target filtering, incorporating clutter density, probabilities of detection, survival and birth of targets \cite{Vo:2005,Mahler:2003,Mahler:2007}. To this end, targets and measurements assume to form sets with variable random cardinalities. Using Finite Set Statistics \cite{Mahler:2003}, the posterior distribution in (\ref{eq:bayes1}) can be extended from vectors to RFS as follows,
\begin{equation}
p_k(X_k|Z_{1:k}) = \frac{g_k(Z_k|X_k)p_{k|k-1}(X_k|Z_{1:k-1})}{\int g_k(Z_k|X)p_{k|k-1}(X|z_{1:k-1})\mu_s(dX)},
\label{eq:bayes2}
\end{equation} 
\noindent where $Z_k$ and $X_k$ are measurement (containing both clutter and true positives) and target RFS, respectively, and $\mu_s$ an appropriate reference measure \cite{Vo:2006}. 
One approach to represent targets is to use Probability Hypothesis Density (PHD) maps \cite{Vo:2005,Vo:2006}. These maps have two basic features: 1) Their peaks correspond to the location of targets; 2) Their integration gives the expected number of targets at each time step. {Vo and Ma proposed Gaussian Mixture PHD (GM-PHD), which propagates the first-order statistical moments to estimate the posterior in (\ref{eq:bayes2}) as a mixture of Gaussians \cite{Vo:2006}.}

\noindent \textbf{{Non-model based approaches:}}
While GM-PHD is based on Gaussian distributions, a particle filter-based solution is proposed by Sequential Monte Carlo PHD (SMC-PHD) to address non-Gaussian distributions \cite{Vo:2003}. 
Since a large number of particles needs to be propagated during SMC-PHD, the computational complexity can be high and hence gating might be necessary.

Cardinalised PHD (CPHD) is proposed by Mahler to also propagate the cardinality of the targets over time in \cite{Mahler:2007}, while its intractability is addressed in \cite{Nagappa:2017}. 
The Labelled Multi-Bernoulli Filter (LMB) is introduced in \cite{Reuter:2014} which performs track-to-track association and outperforms previous algorithms in the sense of not relying on high signal to noise ratio (SNR). 
Vo \emph{et al.} proposed Generalized Labelled Multi-Bernoulli (GLMB) as a labelled multi-target filtering \cite{Vo:2014}. 

\noindent \textbf{{RNNs and LSTM networks:}}
RNNs are neural networks with feedback loops, through which past information can be stored and exploited. They offer promising solutions to difficult tasks such as system identification, prediction, pattern classification, and stochastic sequence modelling \cite{Bay2016}.
Unfortunately, RNNs are known to be particularly hard to train, especially when long temporal dependencies are involved, due to the so-called vanishing gradient phenomenon. Many attempts were carried on in order to address this problem from choosing an appropriate initial configuration of the weights to exploiting orthogonality in the hidden-to-hidden weight matrix \cite{vorontsov2017orthogonality}. Also modifications in the architecture \cite{hochreiter1997} were proposed to sidestep this problem through gating mechanism, which enhance the memory of the network.

The latter case includes LSTMs (which is the architecture we consider in this paper as well). Formally, an LSTM is defined by four gates (input, candidate, forget and output, respectively), i.e. \cite{hochreiter1997}
\begin{eqnarray}
&i_k = \phi_i \bigg( A_i x_k + B_i h_{k-1} + b_i \bigg), \qquad
&j_k = \phi_j \bigg( A_j x_k + B_j h_{k-1} + b_j \bigg) \\
&f_k = \phi_f \bigg( A_f x_k + B_f h_{k-1} + b_f \bigg), \qquad
&o_k = \phi_o \bigg( A_o x_k + B_o h_{k-1} + b_o \bigg),
\end{eqnarray}
for each time step $k$, where $\phi_{\bullet}$ represent element-wise non-linear activation functions {(with $\bullet = i,j,f,o$)}, $A_{\bullet},B_{\bullet},b_{\bullet}$ are the learning weights matrices and bias, $x_k$ is the input, and $h_{k-1}$ is the hidden state at the previous time step. These gates are then combined to update the memory cell unit and compute the new hidden state as follows \cite{hochreiter1997},
\begin{align}
c_k &= c_{k-1} \odot f_k + i_k \odot j_k, \qquad
h_k = \tanh(c_k) \odot o_k,
\end{align}
where $\odot$ represent the element-wise product.
\\Finally, the hidden state is mapped through a fully-connected layer to estimate the predicted output $y_k$ \cite{hochreiter1997},
\begin{equation}\label{eq:outputLayer}
y_k = \phi_y (C h_k + b_y ),
\end{equation}
where similarly to the RNN equation, $\phi_y$ is the element-wise output function and $C$ and $b_y$ are learning weight matrix and bias vector, respectively.

\section{Overall pipeline}\label{sec:overall}
\begin{figure*}[t]
\centering
\includegraphics[width=1.0\textwidth]{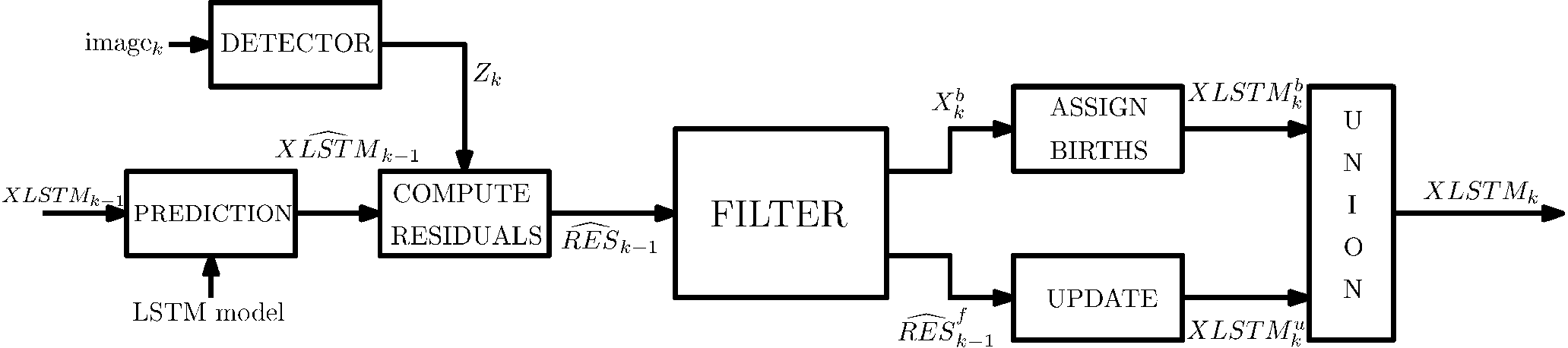}
\caption{Overall pipeline of the proposed multi-target filtering algorithm: target tuples set $XLSTM_{k-1}$ is given to the predictor, which loads the LSTM model for each target, trains it over the target's latest data patch and predicts the target state set $\hat{XLSTM}_{k-1}$. The measurement set $Z_k$ is obtained from the detector, which is used to compute the residual set. The filtering, data association and update (correction) steps are then performed over the residual set to remove clutter from the data and assign births, which eventually, form the target tuple set $XLSTM_k$.}
\label{fig:overall}
\end{figure*}

The overall pipeline is shown in Fig.~\ref{fig:overall} in a block diagram. First, the predicted locations of the target tuples from the previous time step are computed. Then given the current measurements, a set of "residuals" are calculated for each target. These residuals are then used to perform filtering (rejecting the false positives and obtain survivals) and birth assignments. The union of the resulting birth and survival tuple sets are finally used as the targets for the $k^{th}$ iteration. In the following sections each of these steps are detailed.

\section{Online motion modelling via LSTM}\label{sec:lstmMerge}


The target state variations over video frames can be seen as a sequential learning problem.
The target tuples are predicted using an LSTM network. After being updated with their associated measurement, training patches are updated for each target which are used to re-train the very same LSTM model. As a result of this recursive process, motion modelling of the incoming data is performed. Using this approach, a non-linear/non-Gaussian input is learned without incorporating any prior knowledge about the motion. 
In our work, we investigated assigning an LSTM network to each target, which is separately trained over the predicted data from the previous time steps in order to predict for the following target position. The main issue with this approach is the (GPU) memory management and its re-usability. When the number of targets increases, it becomes infeasible to release the (GPU) memory part related to the absent targets and re-allocate the memory to the new targets.

One solution to this is to define a single LSTM network as a graph, whose nodes are simply "placeholders" \cite{Abadi:2016} pointer variables. Every time a target is present in the scene, the placeholder nodes are updated with those weights and biases corresponding to this target. Once the prediction is performed, the memory is released. 
Such memory allocation enables learning a {\emph{global target motion}} over the video sequence, which can be useful in crowd behaviour detection, but will provide poor results analysing the targets separately.

Thus, as a solution we propose an online LSTM training, where each target shares the same LSTM weights (i.e.\ $A_{\bullet},B_{\bullet},b_{\bullet},C,b_y$) with the other targets. During the online training step, these weights and biases are shared from other targets and fine-tuned based on the past measurements.
In other words, we fine-tune the LSTM weights and biases for each target by transferring the learned weights and biases from another targets.
This gives us the advantage to save memory storage even further, gradually reduce the re-training number of epochs, and have predictions which are mainly affected by the recent information for each specific target. In the following sections, we describe the filtering pipeline and show how the data is prepared to be given to the LSTM network for the training and prediction steps.

\section{Tracklet tuples and data association}\label{sec:DataAs}
We define a tracklet tuple $xlstm$ as a subset of $xlstm \in XLSTM_k$ at time $k$, containing the following four components,
\begin{equation}
xlstm = \left(x \in X_k, m_k \in \mathbb{Z}, g_k \in \mathbb{R^{+}}, f_k \in \{0, 1\}\right).
\label{eq:xl}
\end{equation}
\noindent $x$ is an $M \times d$ target state matrix over a $d$-dimensional space. $M$ is the number of previous prediction appended for this target, which are used as the training patch for the LSTM network. $m_k$ is an integer indicating the target {\emph{maturity}} until the $k^{th}$ iteration. $g_k$ is a real positive number containing the target {\emph{genuinity}} error. $f_k$ is a binary {\emph{freeze}} state variable, which is {either} 1 when an observed measurement is used to update the state of the target, {or} is 0 when the update is performed without any measurement but with the target's past states. $XLSTM_k$ and $X_k$ are random finite sets with $M_k$ cardinalities, containing all the target tuples at $k$, i.e $xlstm \in XLSTM_k$ and $x \in X_k$.

\subsection{Filtering and birth assignment}
The LSTM architecture explained in Section~\ref{sec:lstmMerge} is used to predict target state $x_{k|k-1}$. This is performed by first sequentially training the LSTM network with its $M$ samples from the $xlstm$ tuple. Then the sample in the last (the $M^{th}$) row of $x$ is given to the trained network to predict $\hat{x}$, which is then appended to the input tuple $xlstm$ to create the predicted tuple $\hat{xlstm} \in XLSTM_{k|k-1}$. The resulting $XLSTM_{k|k-1}$ is hence an RFS with $M_{k-1}$ cardinality, similar to $XLSTM_{k-1}$. 
At the $k^{th}$ time step, a set of residuals are calculated using the obtained measurement RFS $Z_k$. If $Z_k$ has $N_k$ cardinality, assuming no gating is performed, there will be $N_k \times M_{k-1}$ residuals which are stored as $RES_{k|k-1}$, where $\hat{res} \in RES_{k|k-1}$ has the following structure,
\begin{equation}
\hat{res} = \left(\hat{xlstm} \in XLSTM_{k|k-1}, T \in \mathbb{R^{+}}, z \in Z_k\right),
\label{eq:myres}
\end{equation}
\noindent in which $T$ is the \emph{targetness} error parameter, which is computed from $T = \left\lVert \hat{x} - z \right\rVert_{2}$. The value of $T$ shows how close the predicted target is to the current measurement. 

\noindent $RES_{k|k-1}$ is used to perform the filtering step, at which survival of targets are determined, new births are assigned and false positive targets and measurements are removed.
To do this, first an $N_k \times M_{k-1}$ targetness matrix $T_{k|k-1}$ is constructed, whose element at the $n_k^{th}$ row and $\hat{m}_k^{th}$ column shows the second norm between the $n_k^{th}$ measurement and $\hat{m}_k^{th}$ predicted target ($n_k = 1, 2, \ldots N_k$ and $\hat{m}_k = 1, 2, \ldots M_{k-1}$). In the next section, we detail how $T_{k|k-1}$ is used to perform data association.

\subsubsection{Data association}
For each column and row of $T_{k|k-1}$ the measurement and target indexes are computed, respectively, as follows,
\begin{equation}
C^{I}_{k|k-1} = \underset{\hat{m}_k}{\mathrm{argmin}} \left(T_{k|k-1}\right), \qquad
R^{I}_{k|k-1} = \underset{n_k}{\mathrm{argmin}} \left(T_{k|k-1}\right),
\end{equation}
\noindent where $C^{I}_{k|k-1}$ and $R^{I}_{k|k-1}$ are $N_k \times 1$ and $1 \times M_{k-1}$ vectors, containing the minimum target and measurement indexes, respectively. In addition, the minimum of each column of $T_{k|k-1}$ is also computed,
\begin{equation}
mg_{k|k-1} = \underset{\hat{m}_k}{\mathrm{min}} \left(T_{k|k-1}\right), \qquad
g_{k|k-1} = \underset{n_k}{\mathrm{min}} \left(T_{k|k-1}\right),
\end{equation} 
\noindent {where} $mg_{k|k-1}$ and $g_{k|k-1}$ are $N_k \times 1$ and $1 \times M_{k-1}$ vectors containing the measurement and target genuinity errors, respectively. In other words, $mg_{k|k-1}(n_k)$ and $g_{k|k-1}\left(\hat{m}_k\right)$ are the measurement and target genuinity errors for the $n_k^{th}$ and $\hat{m}_k^{th}$ measurement and target, respectively.
During the next step, the histogram bin of both $C^{I}_{k|k-1}$ and $R^{I}_{k|k-1}$ are computed as $H^{C^{I}}_{k|k-1}$ and $H^{R^{I}}_{k|k-1}$, respectively as follows,
\begin{equation}
H^{C^{I}}_{k|k-1} = \mathrm{hist}(C^{I}_{k|k-1}), \qquad
H^{R^{I}}_{k|k-1} = \mathrm{hist}(R^{I}_{k|k-1}).
\end{equation}
\noindent Each element of $H^{C^{I}}_{k|k-1}\left(\hat{m}_k\right)$ shows how many associations exist for the $\hat{m}_k$ predicted target. On the other hand, each element of $H^{R^{I}}_{k|k-1}\left(n_k\right)$ indicates the number of association to the $n_k^{th}$ measurement. The states of the targets are then updated using the data association approach explained as a pseudo code in Algorithm \ref{Al:dataAssociation}.

\begin{algorithm}[t]
\footnotesize
{
\KwIn{$m_{min}, g_{min}, g_{max}, H^{C^{I}}_{k|k-1}, H^{R^{I}}_{k|k-1}, g_{k|k-1}, mg_{k|k-1},$\\ $XLSTM_{k|k-1}, Z_k$}
\KwOut{Survived targets and births: $RES^{f}_{k|k-1}$ and $X_k^b$}
\% Iterate over $M_{k-1}$ targets in $XLSTM_{k|k-1}$:\\
\For{$\hat{m}_k = 1, 2, \ldots, M_{k-1}$}{
\If{($H^{C^{I}}_{k|k-1}(\hat{m}_k) == 0$ AND $\hat{m}_k^{th}$\textrm{'s} \textrm{target maturity} $\geq m_{min}$) OR ($H^{C^{I}}_{k|k-1}(\hat{m}_k) \geq 1$ AND $g_{min} \leq g_{k|k-1}(\hat{m}_k) \leq g_{max}$)}{
Possible occluded target or association with clutter:
Freeze and decrement maturity\;
}
\If{$H^{C^{I}}_{k|k-1}(\hat{m}_k) \geq 1$ AND $g_{k|k-1}(\hat{m}_k) < g_{min}$}{
Possible target survival:
Unfreeze, increment maturity and associate the target with the $R^{I}_{k|k-1}(\hat{m}_k)^{th}$ measurement in $Z_k$\;
}
}
\% Iterate over $N_{k}$ measurements in $Z_k$:\\
\For{$n_k = 1, 2, \ldots, N_{k}$}{
\If{$H^{R^{I}}_{k|k-1}(n_k) == 0$ OR $mg_{k|k-1}(n_k) > g_{max}$}{
Possible birth of a target:
Initialise a new $xlstm$ tuple with $m_{min}$ maturity\;
}
}
}
\caption{Data association algorithm.}
\label{Al:dataAssociation}
\end{algorithm}

\subsection{Update survivals and assign births}
The survived targets form an RFS $RES^{f}_{k|k-1}$. A set member $res^f \in RES^{f}_{k|k-1}$ has the same structure as $\hat{res}$ in (\ref{eq:myres}), with the only difference that its $m$, $g$, $f$ and $z$ are being updated according to the cases explained in Algorithm \ref{Al:dataAssociation}. During the update stage (shown in the block diagram of Fig.~\ref{fig:overall}), if the freeze state of a target is zero, meaning that a measurement $z_k$ has successfully been associated with the target, its $x_k$ is being updated by appending the associated measurement $z_k$, i.e, $x_k = \left[\begin{smallmatrix}
x_{k-1}\\
--\\
z_k
\end{smallmatrix}\right]$. On the other hand, if the freeze state is one, meaning that the association step failed to find a measurement for the current target (possibly due to occlusion, measurement failure or the target itself is a false positive), the predicted target state $\hat{x}_k$ is appended to $x_{k-1}$ to create the new state matrix $x_k = \left[\begin{smallmatrix}
x_{k-1}\\
--\\
\hat{x}_k
\end{smallmatrix}\right]$.
For both cases, to optimise memory allocation we define a maximum batch size. If the number of rows in $x_k$ was greater than the batch size, the first row of $x_k$ which corresponds to the oldest saved prediction or measurement is being removed.
Using the updated target states, an RFS $XLSTM^{u}_k$ is generated for the survived targets. Each of its members ($xlstm^{u} \in XLSTM^{u}_k$) is a tuple, having the same structure as (\ref{eq:xl}), with updated states computed according to the data association.

In parallel with the above procedure, for each $1 \times d$ birth vector $x^b$, a target tuple is assigned as,
$xlstm^b \in XLSTM^b_{k} = \left(x^b, m_{init}, g = 0, f = 0 \right)$.
The target tuple at the $k^{th}$ time step is calculated as the union of births and survivals, i.e.\ $XLSTM_k = XLSTM^b_{k} \cup XLSTM^u_{k}$, which has $M_k$ cardinality.

\section{Experimental results}\label{sec:expRes}
In this section we present the experimental results of our method in a controlled simulation on synthetic multi-target data.
We compute the Optimal Sub-Pattern Assignment (OSPA, \cite{Schuhmacher:2008}) distance, an improved version of the Optimal Mass Transfer (OMAT) metric to quantitatively evaluate the proposed algorithm. Assuming two sets $A=\{a_1,a_2,\ldots,a_\alpha\}$ and $B=\{b_1,b_2,\ldots,b_\beta\}$, the OSPA distance of order $p$ and cut-off $c$ is defined as \cite{Schuhmacher:2008},
\begin{equation}
\textrm{OSPA}(A,B) = \frac{1}{\max\{\alpha,\beta\}} \bigg( c^p |\alpha-\beta| + cost \bigg)^{1/p},
\label{eq:OSPA}
\end{equation}
where $\alpha$ and $\beta$ are the number of elements in $A$ and $B$, respectively. It essentially consists of two terms: one is related to the difference in the number of elements in the sets $X$ and $Y$ (cardinality error); and the other related to the localisation $cost$ (Loc), which is the smallest pair-wise distance among all the elements in the two sets (the best-worst objective function \cite{Emambakhsh:2017}). In our work, we have used the Hungarian assignment to compute this minimal distance. As in \cite{Schuhmacher:2008}, we choose $p=1$ and $c=100$. OSPA has been widely used for evaluating the accuracy of the filtering and tracking algorithms~\cite{Vo:2017,Fantacci:2018}. 
The overall pipeline is implemented (end-to-end) in Python 2.7, and all the experiments are tested using an NVIDIA GeForce GTX 1080 GPU and an $i5-8400$ CPU. We have used a 3-layer LSTM network, each having 20 hidden units, outputting a fully-connected layer (\ref{eq:outputLayer}), with $\phi_y$ as an identity function. 
The network is trained online at each time step for 50 epochs over the currently updated patch for each target, minimising the mean square error as the loss function and using Adam optimisation method \cite{kingma2014adam}.

In order to evaluate our algorithm over different filtering problems, such as occlusions, birth and death of the targets, non-linear motion and spawn, we have used the multi-target simulation introduced by Vo and Ma~\cite{Reuter:2014,Vo:2014}. In this scenario, there are 10 targets appearing in the scene having various birth times and lifespans. The measurements is performed by computing the range and bearing (azimuth) of a target from the origin. It also contains clutter with uniform distribution along range and azimuth, with a random intensity sampled from a Poisson distribution with $\lambda_c$ mean. The obtained measurements are degraded by a Gaussian noise with zero mean and $\sigma_r=10$ (unit distance) and $\sigma_\theta=\pi/90$ (rad) standard deviation, respectively. The problem is to perform online multi-target filtering to recover true positives from clutter.
In our first experiment we compute the OSPA error, assuming $\lambda_c = 20$ clutter intensity. 


\begin{table}[t]
\small{
\begin{center}
\begin{tabular}{c | c c c} 
Algorithm & OSPA Card & OSPA Loc & OSPA \\
\hline
PHD-EKF & $9.25 \pm 15.44$ & $20.86 \pm 9.83$ & $30.11 \pm 14.82$ \\
PHD-SMC & $12.76 \pm 15.67$ & $46.08 \pm 17.36$ & $58.84 \pm 15.85$ \\
PHD-UKF & $10.33 \pm 18.17$ & $19.73 \pm 8.14$ & $30.06 \pm 16.21$ \\
\hdashline
CPHD-EKF & $7.10 \pm 14.91$ & $23.00 \pm 11.06$ & $30.10 \pm 16.16$ \\
CPHD-SMC & $11.18 \pm 13.72$ & $46.08 \pm 18.59$ & $57.25 \pm 17.72$ \\
CPHD-UKF & $5.50 \pm 14.79$ & $22.39 \pm 11.21$ & $27.89 \pm 15.43$ \\
\hdashline
LMB-EKF & $4.59 \pm 14.60$ & $22.59 \pm 8.60$ & $27.18 \pm 13.96$ \\
LMB-SMC & $12.07 \pm 19.75$ & $23.47 \pm 13.49$ & $35.54 \pm 17.95$ \\
LMB-UKF & $\textbf{3.77} \pm \textbf{13.82}$ & $21.94 \pm 10.22$ & $25.72 \pm 14.39$ \\
\hdashline
GLMB-EKF & $6.37 \pm 17.66$ & $20.13 \pm 8.02$ & $26.50 \pm 15.20$ \\
GLMB-SMC & $6.11 \pm 11.91$ & $21.07 \pm 6.78$ & $27.19 \pm 11.49$ \\
GLMB-UKF & $11.79 \pm 16.34$ & $19.84 \pm 9.75$ & $31.63 \pm 15.12$ \\
\hdashline
\textbf{OURS} & $10.36 \pm 13.25$ & $\textbf{8.77} \pm \textbf{7.50}$ & $\textbf{19.12} \pm \textbf{14.39}$ \\
\end{tabular}
\end{center}
}

\caption{OSPA error for different methods: we compared our approach to PHD, CPHD \cite{Nagappa:2017,Mahler:2007}, LMB \cite{Reuter:2014}, and GLMB \cite{Vo:2014,Vo:2017} algorithm, when EKF, SMC, and UKF are used for prediction and update steps. The best performer method is highlighted in bold.}
\label{tab:table1}
\end{table}

\begin{figure}[!b]
\centering
\includegraphics[width=0.5\textwidth]{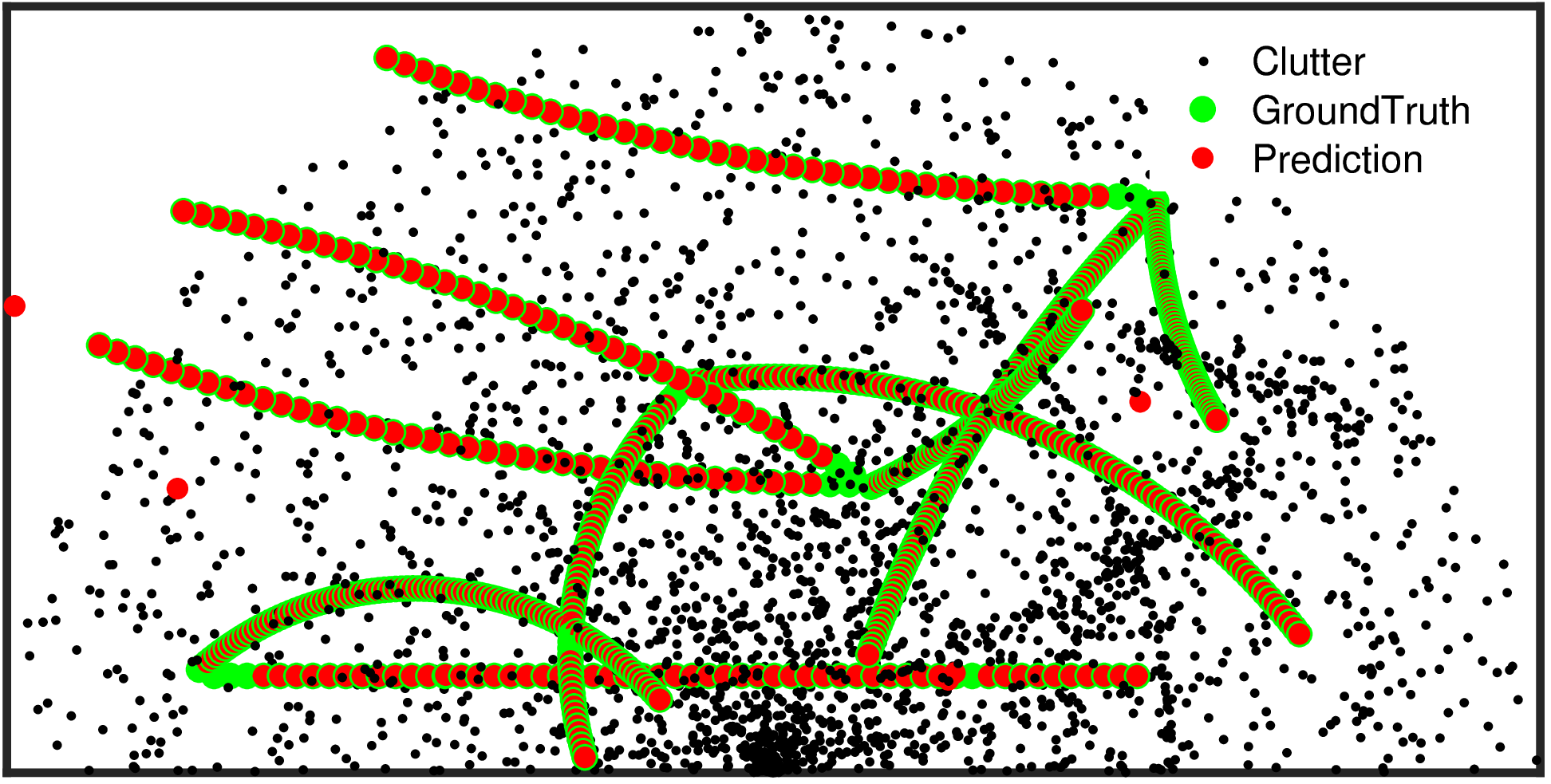}
\caption{Results of trajectories for the ten simulated points.}
\label{fig:SimEnvRes}
\end{figure}

In Table \ref{tab:table1}, we report the average and standard deviation for the overall OSPA (see \eqref{eq:OSPA}) and its two terms related to cardinality error (OSPA Card) and optimal Hungarian distance (OSPA Loc - the $cost$ term in (\ref{eq:OSPA})). We compare our method with PHD, CPHD, LMB, and GLMB algorithm, when EKF, SMC, and UKF used as basis for the prediction and update steps (The following Matlab implementation of these algorithms is used: \url{http://ba-tuong.vo-au.com/codes.html}). Our method outperform all the other algorithms in terms of overall OSPA. In particular, this is due to a significant drop of the Loc error, while cardinality error is comparable with most of the others. For example, despite our algorithm has $\approx 7$ (samples) higher average OSPA cardinality error than LMB-UKF, our Loc and overall OSPA distances are about $\approx 13$ and $6$ (unit distances) lower, respectively.
The resulting trajectories of this experiment for our method are illustrated in Fig.~\ref{fig:SimEnvRes}. The red dots represent the predicted location of the targets at every time step, filtered out from the measurements clutters (black dots). They almost overlap with the ground truth (green dots), except very few (only three) false positives (predicted but no ground truth) and false negatives (ground truth but no prediction).

\begin{figure}[!t]
\centering
\includegraphics[width=0.65\textwidth]{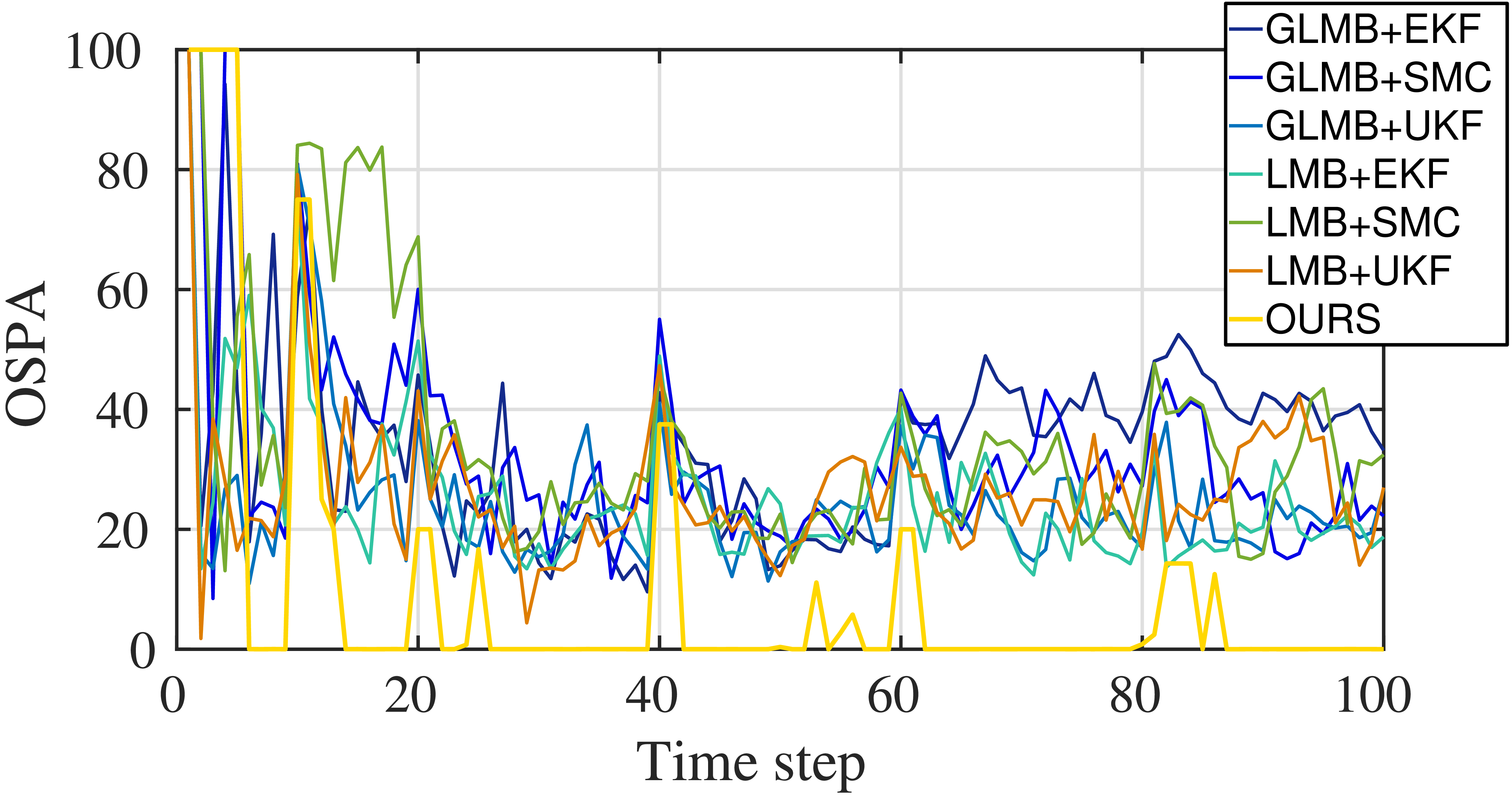}
\caption{Comparison of overall OSPA error for different methods for $\lambda_c=20$. Except the very early time steps when the LSTM have not yet learned the motion, our method has a remarkable better performance than the other filters.}
\label{fig:OSPA_time}
\end{figure}

\begin{figure}[!t]
\centering
\includegraphics[width=0.65\textwidth]{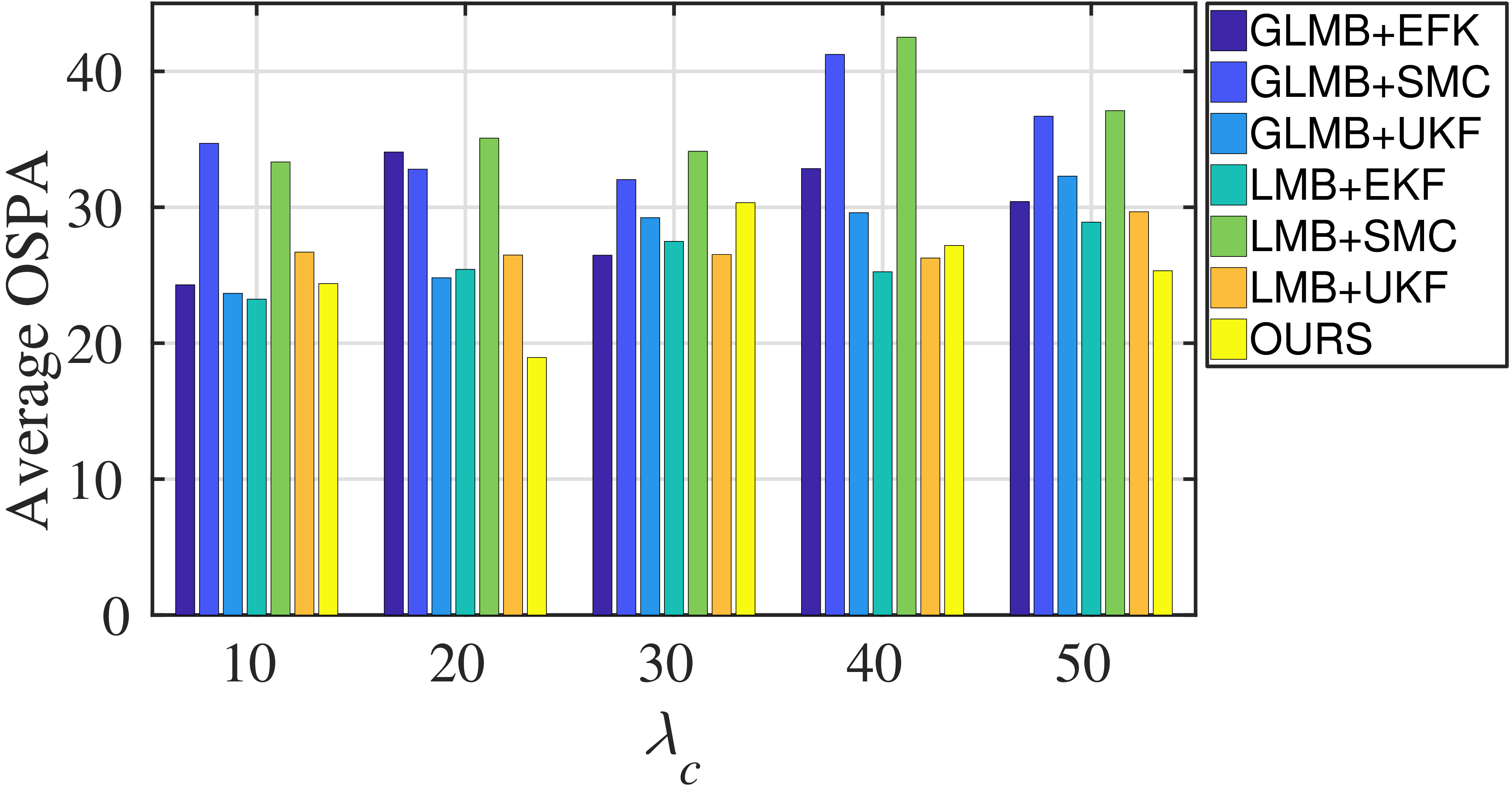}
\caption{Overall OSPA for different methods, using different clutter intensity $\lambda_c$.}
\label{fig:OSPAClutter}
\end{figure}

Moreover, in Fig.~\ref{fig:OSPA_time}, we show the overall OSPA at every time steps. During the initial time steps (frame number $ < 6$), our OSPA error is higher. This is mostly due to the under-fitting of the LSTM model because of lack of data. However, after $\approx 7^{th}$ iteration our OSPA error becomes significantly lower than other approaches, having an overall average of $\approx 18$, while the average OSPA error for other algorithms are $> 25$. 
In order to show the robustness of our algorithm for higher clutter densities, in the second experiment, we increase the clutter intensity $\lambda_c$ and find the average OSPA over all time steps. Figure~\ref{fig:OSPAClutter} shows the results of this experiment, for $\lambda_c = 10, 20, \ldots, 50$. Our filtering algorithm provides a relatively constant and comparably lower overall OSPA error even when the clutter is increased to 50. Both of the SMC-based algorithms (GLMB-SMC and LMB-SMC) generate highest OSPA error, which can be due to the particle filter algorithm divergence. On the other hand, lower OSPA errors generated by the LMB with an EKF model shows how successfully this particular simulated scenario can be modelled using such non-linear filter. It should be mentioned, however, that our method does not rely on any prior motion model capable of learning the non-linearity within the data sequence.

\section{Conclusions}\label{sec:conc}
This paper addressed the problem of fixed motion and measurement models for the multi-target filtering using an adaptive deep learning framework. This is performed by defining target tuples with random finite set terminology and utilisation of LSTM networks, learning to model the target motion while simultaneously filtering the clutter. We defined a novel data association algorithm compatible with the predicted tracklet tuples, enabling the update of occluded targets, in addition to assigning birth, survival and death of targets. Finally, the algorithm is evaluated over a commonly used filtering scenario via OSPA metric computation.

Our algorithm can be extended by investigating an end-to-end solution for tracking, encapsulating the data association step within the recurrent neural network architecture.


\bibliographystyle{splncs04}
\bibliography{refs}
\end{document}